\documentclass[lettersize,journal]{IEEEtran}
\usepackage{amsmath,amsfonts}
\usepackage{array}
\usepackage[caption=false,font=normalsize,labelfont=sf,textfont=sf]{subfig}
\usepackage{textcomp}
\usepackage{stfloats}
\usepackage{url}
\usepackage{verbatim}
\usepackage{graphicx}
\usepackage{cite}
\usepackage{xcolor}
\usepackage{hyperref}
\usepackage{url}
\usepackage{here}
\usepackage{algorithm}
\usepackage{amssymb} 
\usepackage[noend]{algpseudocode}
\algrenewcommand\algorithmicrequire{\textbf{Initiate}}
\usepackage{tabularx}
\usepackage[bottom]{footmisc}
\usepackage{comment}
\usepackage{nomencl}

\makenomenclature

\begin{document}

\title{Human-in-the-Loop Pareto Optimization{:}\\  Trade-off Characterization {for} Assist-as-Needed Training and Performance Evaluation}
\author{Harun Tolasa, \IEEEmembership{Student Member,~IEEE} \hspace{25mm} Volkan Patoglu, \IEEEmembership{Member,~IEEE} \hspace{7mm} 
\thanks{H. Tolasa and V. Patoglu are with the Faculty of Engineering and Natural Sciences at Sabanci University, Istanbul, Turkiye. {\tt\small \{harun.tolasa, volkan.patoglu\}@sabanciuniv.edu}. This work has been partially supported by TUBITAK Grants~120N523 and~23AG003.}}

\maketitle

\begin{abstract}
During human motor skill training and physical rehabilitation, there is an inherent trade-off between task difficulty and user performance. Characterizing this trade-off is crucial for evaluating user performance, designing assist-as-needed~(AAN) protocols, and assessing the efficacy of training protocols. In this study, we propose a novel \emph{human-in-the-loop}~(HiL) \emph{Pareto optimization} approach to characterize the trade-off between task performance and the \emph{perceived} challenge level of motor learning or rehabilitation tasks. We adapt \emph{Bayesian multi-criteria optimization} to systematically and efficiently perform HiL Pareto characterizations. Our HiL optimization employs a hybrid model that measures performance with a quantitative metric, while the perceived challenge level is captured with a qualitative metric derived from preference-based user feedback. We demonstrate the feasibility of the proposed HiL Pareto characterization through a user study. Furthermore, we present the utility of the framework through three use cases in the context of a manual skill training task administered to healthy individuals with haptic feedback. First, we demonstrate how the characterized trade-off can be used to design a sample AAN training protocol for a motor learning task and to evaluate the group-level efficacy of the proposed AAN protocol relative to a baseline adaptive assistance protocol. Second, we demonstrate that individual-level comparisons of the trade-offs characterized before and after the training session enable fair evaluation of training progress under different assistance levels. This evaluation method is more general than standard performance evaluations, as it can provide insights even when users cannot perform the task without assistance. Third, we show that the characterized trade-offs also enable fair performance comparisons among different users, as they capture the best possible performance of each user under all feasible assistance~levels.
 \end{abstract}


\IEEEpeerreviewmaketitle

\section{Introduction}~\label{Introduction}
\vspace{-3mm}

Physical human-robot interaction~(pHRI) is commonly used for training tasks, such as human motor skill training or robot-assisted physical rehabilitation. In such applications, the efficacy of the training, typically measured by performance evaluations administered after the training without any assistance, is of utmost importance. Haptic assistance is provided only during the training sessions, when the user is coupled to a force-feedback robot, to ensure that the execution of the task can be completed with sufficient success. For instance, consider a stroke patient going through physical rehabilitation, for whom task execution may not be feasible without a proper level of assistance. For such a patient, some level of assistance is necessary for task completion. On the other hand, too much assistance during training is known to be detrimental, as users may learn to rely on the existence of this support and slack~\cite{Li2009a, Erdogan2012}. Too much assistance may also cause the task to be perceived as not sufficiently challenging by the patient, resulting in a lack of engagement. As a consequence of excessive assistance, users may not learn how to successfully execute the task when no assistance is available. Accordingly, there exists an inherent trade-off between the level of assistance provided and the level of challenge perceived by the user for any given motor control/rehabilitation task. 

This trade-off has been widely acknowledged in the literature, and it has been established that an optimal assistance level consists of the least assistance that allows a user to execute a task with a sufficient level of success. Controllers that aim to provide such assistance are commonly referred to as assist-as-needed~(AAN) controllers. AAN controllers aim to keep the assistance level at a proper level to maximally challenge but not to overwhelm users with task difficulty or demotivate them with continual failures.

The literature on AAN control mostly focuses on the design of interaction controllers that can administer assistance forces safely and naturally, without overriding users' intent. In particular, path-following controllers, such as velocity field controllers~\cite{Moreno2007} and controllers with guaranteed coupled stability properties, such as passive velocity field controllers~\cite{Li1999}, have been proposed and adopted in many related works~\cite{Colombo2011,Erdogan2011c, Keller2013,Asl2020,Lopes2020}. Although these studies on interaction control are indispensable for the safe and natural delivery of force feedback, these control approaches necessitate the proper assistance level to be provided as an input, typically by a domain expert. 

In AAN controllers, the proper level of assistance is commonly decided empirically or heuristically, based on thresholds imposed on the measured/estimated performance of the user. Most methods utilize sensor inputs or biosignals~\cite{Krebs2003,Yanfang2009,Mine2013,Ozdenizci_2017,Yang2023}, such as EMG and EEG, to adjust the level of assistance to promote voluntary participation based on some pre-determined thresholds. Adaptive controllers that utilize the dynamic model of the user and the device to minimize a cost function~\cite{Emken2005} or rely on statistical estimates of psychophysical thresholds~\cite{Squeri} have also been proposed. Reinforcement learning algorithms have been used to develop AAN controllers that do not depend on user- or device-specific parameters~\cite{Pareek2023}. Furthermore, in addition to quantitative measures, more qualitative aspects, such as the psychological states of the user, have been estimated by neural networks to adjust assistance levels~\cite{Zhong2019,Koenig2011}. Interested readers are referred to the review articles~\cite{Baud2021,Mahfouz2024} for recent implementations of AAN control techniques for lower- and upper-extremity rehabilitation.

While these studies are quite promising, the characterization of the underlying trade-off between the performance versus the perceived task difficulty and the determination of a customized level of assistance to be provided to a user are still open challenges, commonly delegated to a domain expert. The problem is challenging as each individual is unique; furthermore, user preferences and perceptions undergo continuous changes as learning/recovery takes place with training. For instance, the perceived challenge level of a task under a certain level of assistance is likely to decrease as a user gets better at the task. User-dependent metrics that are not easy to quantify, such as motivation and comfort level, may directly affect the perceived challenge level. Hence, determining the ideal assistance level necessitates personalization, possibly through a characterization of the underlying trade-off between the perceived challenge level and the performance for each user.

\begin{figure*}[t!]
\centering \vspace{-0.25\baselineskip}
\includegraphics[width=.825\textwidth]{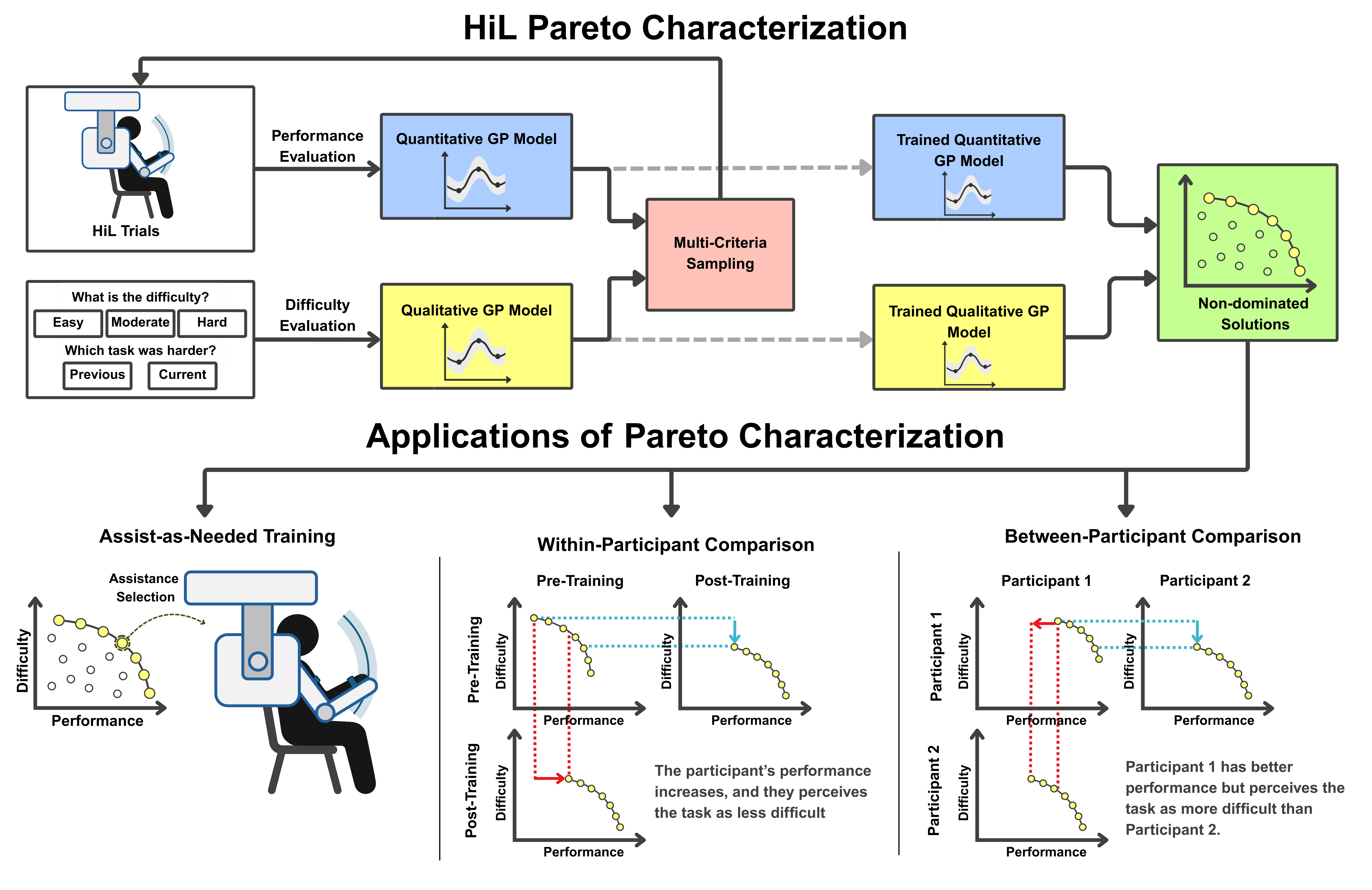}
\vspace{-1.5\baselineskip}
\caption{{Proposed HiL Pareto characterization scheme and its application to AAN training, within- and between-participant performance evaluations.}} 
\label{fig:HiLParetoBigFigure}
\end{figure*}

Characterization of the trade-off between the perceived challenge level and the task performance can guide training by helping designers to establish a proper level of assistance for AAN controllers. Moreover, such characterizations can be performed at different stages of the training, such that changes in user preferences/performance can be captured as learning takes place and the assistance can be adjusted accordingly.

Furthermore, such trade-off characterizations can serve as a novel means for evaluating performance at a group- or individual-level. While typical evaluations of manual skill training or physical rehabilitation are performed when no assistance is provided, as this represents the real-life use case, such evaluations cannot capture improvements in performance during the early phases of training/rehabilitation. For instance, if a task is excessively hard for a patient, then it may not be possible for the patient to execute it successfully without a proper level of assistance. In such cases, evaluations with no assistance will not capture any improvement with training, as the patient's success rate will remain low. 

On the other hand, a comparison of the characterized trade-offs between the perceived challenge level versus the performance of a user, or a group of users, at different stages of training provides a feasible alternative: the time evaluation of the trade-off provides a rigorous means to fairly assess the progress, under all feasible assistance~levels. 

In addition, such trade-off characterizations also enable comparisons among various users or groups of users. It is important to emphasize that ensuring the rigor and fairness of such performance comparisons under assistance necessitates the characterization of the trade-off via Pareto optimization, since multiple variables, such as assistance and the perceived challenge levels, need to be considered simultaneously. 

In this study, we propose a human-in-the-loop~(HiL) \emph{Pareto} optimization approach to characterize the trade-off (i.e., Pareto optimal results forming a set of non-dominated solutions) between the user's performance and the perceived challenge level for a motor learning task. During the optimization, the user performance is measured by a quantitative metric, while the perceived challenge level is captured by a qualitative metric gathered through preference-based qualitative feedback, as depicted in Figure~\ref{fig:HiLParetoBigFigure}. A multi-criteria Bayesian optimization technique is utilized for HiL Pareto optimization of this hybrid model with quantitative and qualitative metrics. The sample-efficiency of the underlying Bayesian optimization is a crucial aspect, as it enables the trade-off to be characterized systematically and efficiently by focusing on the relevant portion of the search space and without inducing fatigue to participants. 

Once the trade-off is characterized via HiL Pareto optimization, first, we demonstrate how this trade-off can be used to guide the training sessions with Pareto optimal assistance levels. In particular, we show how a subset of optimal solutions can be selected from the set of non-dominated solutions to guide AAN training sessions. Second, we show that the trade-off evolves in time as learning takes place, and Pareto-front curves capturing the trade-off can be used to fairly evaluate both the individual- and group-level progress. Third, we show that the characterized trade-offs can also be used for fair performance comparisons among \emph{different users}, by capturing the best possible performance of each user under all feasible assistance levels. \vspace{-1.5mm}

\subsection*{Contributions}

The main contribution of this study is a novel HiL Pareto optimization framework with hybrid (quantitative and qualitative) performance measures, as depicted in the first row of Figure~\ref{fig:HiLParetoBigFigure}. We not only formulate the HiL trade-off characterization framework but also show its feasibility through a user study by demonstrating that the proposed HiL Pareto optimization can be used to systematically and efficiently characterize the trade-off between the performance and the perceived challenge level of a motor skill learning task.
 
Our second contribution is the demonstration of the usefulness of the HiL trade-off characterization through three use cases, as depicted in the second row of Figure~\ref{fig:HiLParetoBigFigure}. Within the context of a case study involving a manual skill training task administered to healthy individuals with haptic feedback, we demonstrate that 

\begin{itemize}
 \item[-] the non-dominated solutions characterizing the trade-off between the performance and the perceived challenge level can be used to design AAN controllers,
 \item[-]  the comparisons of the trade-off curves characterized at different stages of training can be used to establish a novel and rigorous means of individual- and group-level performance evaluation across assistance levels, and
 \item[-] the trade-off curves of different users can be used for fair comparisons among various users, as they capture the best possible performance of each user under all feasible assistance levels.
\end{itemize}

\section{Related Work}

Bayesian optimization is a sample-efficient derivative-free global optimization approach~\cite{Brochu2010} that has been employed in several studies involving HiL optimization.  For instance, HiL applications of these algorithms have been used to optimize wearable robotic assistive devices, where the evaluation of optimization metrics is costly, or the number of experiments is constrained by human involvement~\cite{Felt2015, JeffreyRKoller2015, Ding2017, Zhang2017}. 
 
HiL Bayesian optimization studies focus on the optimization of a single-criterion cost function. These techniques are mostly based on quantitative metrics, such as metabolic cost~\cite{Felt2015, JeffreyRKoller2015, Ding2017, Zhang2017}. Recently, qualitative metrics, such as perceived realism, have been captured by probabilistic latent functions and used to implement HiL Bayesian optimizations~\cite{Sadigh2017, Li2020, Biyik2020, Catkin2023, Tolasa2024}.

Bayesian optimization approaches have been extended to solve multi-criteria optimization problems~\cite{Mathern2021, Suzuki2020, Paria2020, Belakaria2020, Emmerich2008, Hernandez2016}. One means to address a multi-criteria optimization problem is to use a weighted sum of the cost functions to form a single aggregate cost function. Such scalarization approaches enable the original multi-objective problem to be formulated as a single-criterion optimization problem. Scalarization approaches have also been applied to the Bayesian optimization setting, either by utilizing an aggregate cost function~\cite{Mathern2021} or a scalarized acquisition function for parameter sampling~\cite{Suzuki2020, Paria2020}. However, in all scalarization methods, since the relative weights of cost functions need to be pre-determined, the design preferences among the objectives must be assigned a priori, before gaining sufficient knowledge of the trade-off~involved. 

On the other hand, for multi-criteria optimization problems, Pareto optimization methods compute a set of non-dominated solutions that correspond to optimal designs for different design preferences among the optimization metrics, and all such \emph{non-dominated solutions} constitute the Pareto-front~\cite{papalambros2000,Marler2010,unalPareto}. Unlike the single-shot scalarization-based optimization methods, Pareto methods fully characterize the trade-off among multiple objectives. Once the Pareto-front solutions are computed, the designer can study these solutions to get an insight into the underlying trade-offs and make an informed decision to finalize the design by selecting a subset of optimal solutions from the Pareto set. 

Pareto-optimization approaches have also been applied in the Bayesian optimization setting~\cite{Belakaria2020,Emmerich2008,Hernandez2016}. 
Existing techniques can be loosely classified as hyper-volume improvement approaches~\cite{Emmerich2008}, information-theoretic methods~\cite{Hernandez2016}, and wrapper methods via single-objective acquisition functions~\cite{Belakaria2020}. In hyper-volume improvement approaches, the 
expected improvement acquisition function is extended for multi-objectives to produce an effective sampling strategy~\cite{Emmerich2008}, while the information-theoretic methods derive a single acquisition function to maximize information gain for all objectives~\cite{Hernandez2016}. Information-theoretic methods have also been extended to work with multiple qualitative metrics~\cite{Astudillo2024}. Finally, the wrapper methods utilize a multi-criteria optimizer to compute a surrogate Pareto-front characterizing the trade-off between the acquisition functions and sample parameters with the highest volumetric variance~\cite{Belakaria2020}. 

The previous studies of the authors focus on \emph{single-criterion} HiL Bayesian optimization with qualitative metrics to improve perceived realism of haptic rendering~\cite{Catkin2023,tolasa2025a} and to explore visual-haptic cue integration during multi-modal haptic rendering under conflicting cues~\cite{Tolasa2024}. This study extends these works to HiL \emph{Pareto} optimization for the characterization of the underlying trade-off between users' performance and perceived challenge level of motor learning tasks. The underlying concepts, mathematical formulations, and optimization approaches used to solve Pareto optimization problems are distinct from those of single-criterion optimization~\cite{papalambros2000}; hence, the extension from single-criterion HiL optimization to HiL Pareto optimization is substantial.
\medskip

Application of Pareto optimization approaches in a HiL setting has only recently been pursued. In addition to this study, a multi-objective HiL optimization problem has recently been addressed in~\cite{Zhang2025} as a case study, utilizing a genetic algorithm-based solver to adjust the assistance profiles of an ankle exoskeleton by simultaneously minimizing two quantitative metrics capturing gait deviations. Our study considers hybrid (both quantitative and qualitative) metrics, relies on a sample-efficient Bayesian Pareto optimization approach, and applies the framework in a human motor skill learning setting. Specifically, our study extends the Bayesian Pareto optimization approach in~\cite{Belakaria2020} to hybrid models and applies it in the~HiL~context for a human motor skill learning task. Our study not only applies HiL Pareto optimization in a motor skill learning setting with hybrid metrics, but also demonstrates its usefulness for the evaluation of individual- and group-level performance and for the design of AAN controllers.

Finally, the comparison of Pareto solutions to enable fair comparisons of performance among various designs is new to the motor skill learning context, while the idea has been successfully applied to comparisons of exoskeleton performance of musculoskeletal simulations~\cite{Bonabarxiv2021} and comparisons of interaction controller performance in the context of~pHRI~\cite{Yusuf2020}.
 
\vspace{-.5mm}
\section{Human-in-the-Loop Pareto Optimization}

The goal of HiL Pareto optimization is to characterize the inherent trade-off between task difficulty and performance. In general, these two metrics are in direct conflict with each other; hence, a multi-objective optimization needs to be performed to consider them simultaneously.

While it is easier to measure task performance directly through quantitative metrics, the task difficulty is much harder to capture, as it depends on many aspects, including but not limited to the sensorimotor skills, as well as engagement, (physical/mental) fatigue, and comfort level of the user. Along these lines, we measure the task performance~($GP_{num}$) quantitatively based on users' scores, as detailed in Section~\ref{Sec:quantitative}, while we capture the perceived challenge level~($GP_{qual}$) of the users by directly acquiring their qualitative feedback, by querying for their ordinal classifications and pair-wise preferences, as detailed in Section~\ref{Sec:qualitative}. Given these two metrics, the following max-max optimization problem is solved over the design variable of assistance percentage, subject to the physical/mental constraints imposed by the user: \vspace{-5mm}


\begin{align*}
    & \max_{\text{assistance \%} \in [0, 100]} \left[ 
    \begin{array}{ll}
        GP_{num}: & \text{task performance} \\ 
        GP_{qual}: & \text{perceived task difficulty}
    \end{array} 
    \right] \\
    & \text{subject to: physical and mental constraints of the user}
\end{align*}\vspace{-5mm}

\begin{algorithm}[b!] 
\caption{HiL Bayesian Pareto Optimization}
\label{Alg:algorithm1}{\small
{ 
\begin{algorithmic}[1]
\Require $A$: Feasible parameter space, 
$GP_{\text{num}}$: Quantitative GP regression model, 
$GP_{\text{qual}}$: Qualitative GP regression model, 
$N$: \# of iterations, 
$N_0$: \# of space-filling iterations
\State Assign acquisition function $\alpha_{\text{num}}$ to $GP_{\text{num}}$
\State Assign acquisition function $\alpha_{\text{qual}}$ to $GP_{\text{qual}}$
\State Create a sampling set $x$ with size $N_0$ using Sobol sequence
\For{$n = 1$ to $N$}
    \If{$n \leq N_0$}
        \State Select $x_n$ from set $x$
    \Else
        \State Compute surrogate Pareto set:
        \State \hspace{1em} $x_p \gets \arg\max_{x \in A} \left( \alpha_{\text{num}}, \alpha_{\text{qual}} \right)$
        \State Select parameter:
        \State \hspace{1em} $x_n \gets \arg\max_{x \in x_p} \left( \sigma_{\text{qual}} \cdot \sigma_{\text{num}} \right)$
        \State Append $x_n$ to set $x$
    \EndIf   
    \State Use $x_n$ in the task trial
    \State Get the performance score $s_n$
    \State Append $s_n$ to $s$ and re-calculate $y$
    \State Get user's qualitative feedback $q_n$
    \State Append $q_n$ to set $q$
    \State Update $GP_{\text{num}}$ using $x$ and standardized score set $y$
    \State Update $GP_{\text{qual}}$ using $x$ and qualitative feedback set $q$
\EndFor
\State Compute Pareto front using the mean predictions of the GP models
\State Plot the Pareto front and list the non-dominated solutions
\end{algorithmic}}}
\end{algorithm}

While performing HiL Pareto optimization, it is important to determine the trade-off among multiple conflicting objective functions, while also minimizing the total resource cost of the experiments.
Among various approaches,  multi-criteria Bayesian optimization techniques hold promise for use in HiL applications due to their inherent sample efficiency~\cite{garnett_bayesoptbook_2023}, as the central idea of all Bayesian optimization approaches is to minimize the number of observations while rapidly converging to the optimal solution(s). Accordingly, Bayesian optimization provides a class of sample-efficient global optimization methods, where a probabilistic model conditioned on previous observations is used to determine the future evaluations. 

The statistical modeling in multi-criteria Bayesian optimization techniques is typically handled by one Gaussian process~(GP) model for each objective to ensure the tractability of the problem, while the selection of the acquisition function to capture the trade-off between multiple objectives results in various alternative techniques~\cite{Suzuki2020,Paria2020,Mathern2021}. 

We utilize a customized version of the wrapper method, called {\emph{Uncertainty-aware Search framework for optimizing Multiple Objectives}}~(USeMO), which is based on single-objective acquisition functions~\cite{Belakaria2020}. USeMO utilizes a multi-criteria sampling strategy that allows one to leverage acquisition functions from single-objective Bayesian optimization to solve the multi-objective  Bayesian optimization problem, as detailed in Section~\ref{Sec:SamplingPareto}. We have preferred USeMO as its application to optimizations with hybrid (qualitative and quantitative) metrics is more accessible. Please note that USeMO is a sample optimization approach appropriate for HiL Pareto optimization, and alternative methods, such as~\cite{Emmerich2008,Hernandez2016}, may also be adapted for HiL Pareto optimization.

\vspace{-3mm}
\subsection{An Overview of HiL Multi-Criteria Bayesian Optimization} \label{multicriteria}

The multi-criteria Bayesian optimization approach, summarized in Algorithm~\ref{Alg:algorithm1}, relies on two GP regression models: one based on quantitative performance scores as detailed in Section~\ref{Sec:quantitative}, and another one based on qualitative feedback collected from the participants as detailed in Section~\ref{Sec:qualitative}. Both GP regression models possess their corresponding acquisition functions.

During the initial iterations, parameters are selected via the Sobol sequence to ensure search-space exploration. For the rest of the iterations, new solutions are computed via the sampling strategy proposed in~\cite{Belakaria2020} and further detailed in Section~\ref{Sec:SamplingPareto}.

In particular, the sampling strategy utilizes a multi-criteria optimizer to compute a surrogate Pareto front characterizing the trade-off between the conflicting acquisition functions. Then, from the set of non-dominated solutions on the surrogate Pareto front, a parameter with the highest volumetric variance is selected for the next sampling. At the end of the trade-off characterization session, Algorithm~\ref{Alg:algorithm1} utilizes a multi-criteria optimizer to compute the non-dominated solutions forming the Pareto front of the expected performance scores and expected perceived challenge level. Once the Pareto set is computed, it can be used for evaluations, or promising non-dominated solutions can be selected as in Section~\ref{Sec:selection} to design AAN training protocols.

\subsection{Quantitative Gaussian Process Regression Model} \label{Sec:quantitative}

A GP regression model, $GP_{num}$, is designed to learn the relationship between the level of assistance and the participant's performance. The maximum applicable assistance level during a task is represented by one, while zero represents the case without assistance. Let $A=\{ x \subset \mathbb{R}^d : 0 \le x_i \le 1\}$ be the feasible parameter space and $x_i$ be the assistance level. Let $\boldsymbol{x}=\{ x_{1},x_{2},..,x_{n}\}$ be a set consisting of $n$ assistance levels used in previous task trials. Then, let $\boldsymbol{s}=\{ s_{1},s_{2},..,s_{n}\}$ be a set consisting of $n$ observed performance scores corresponding to $\boldsymbol{x}$, and let $\boldsymbol{y}=\{ y_{1},y_{2},..,y_{n}\}$ be the statistically standardized version of $\boldsymbol{s}$. The dataset used to train the numerical GP regression model is represented with $\boldsymbol{D_{\mathcal{N}}}=\{(x_1,y_1),(x_2,y_2),...,(x_n,y_n)\}$.

Finally, let $f_{\mathcal{N}}{(x)}$ be the black-box function representing the relationship between the assistance level and the standardized performance scores. To model the deviation of the score measurements, we assume that the standardized performance scores are affected by a Gaussian white noise with variance $\sigma_{w}^2$, and observe their noisy version $y$.
Then, the prior black-box function $f_{\mathcal{N}}$ for the performance scores can be modeled as \vspace{-4mm}

\begin{equation}\label{prior f PDF Eq}
f_{\mathcal{N}}(\boldsymbol{x}) \sim \mathcal{GP}(0,K_{\mathcal{N}}+\sigma_{w}^2I)
\end{equation}

\vspace{-1mm}

\noindent with $K_{\mathcal{N}}\in \mathbb{R}^{n\times n},k_{\mathcal{N}_{i,j}}= k_{\mathcal{N}}(x_i,x_j)$, is a kernel function used for the correlation.  

The performance scores for an unknown assistance level can be predicted via Bayesian inference. Let $x_*$ be any arbitrary assistance level, and let $f_{\mathcal{N}_*|\boldsymbol{D_\mathcal{N}}}$ be the corresponding performance score estimation based on previously acquired performance scores. Then, the Bayesian inference indicates \vspace{-3mm}

\small
\begin{eqnarray}\label{post_GP}
f_{\mathcal{N}_*|\boldsymbol{D_\mathcal{N}}} \!\!\!\!&\sim& \!\!\!\!\mathcal{GP}(\mu_{\mathcal{N}_*|\boldsymbol{D_\mathcal{N}}},\sigma_{\mathcal{N}_*|\boldsymbol{D_\mathcal{N}}}^2) \\
\mu_{\mathcal{N}_*|\boldsymbol{D_\mathcal{N}}} \!\!\!\!&=&\!\!\!\! k_{\mathcal{N}_{*,1:n}}(K_{\mathcal{N}}+\sigma_{w}^2I)^{-1}\boldsymbol{y}\\
\sigma_{\mathcal{N}_*|\boldsymbol{D_\mathcal{N}}}^2 \!\!\!\!&=&\!\!\!\! k_{\mathcal{N}_{*,*}}-k_{\mathcal{N}_{*,1:n}}(K_{\mathcal{N}}+\sigma_{w}^2I)^{-1}k_{\mathcal{N}_{1:n,*}}
\end{eqnarray} \normalsize

\vspace{-3mm}
\subsection{Qualitative Gaussian Process Regression Model} \label{Sec:qualitative}

A second GP regression model, $GP_{qual}$, is designed to learn the relationship between the assistance level and the perceived challenge of the task as perceived by the participant. During the Pareto characterization session, after each task trial, the participant classifies the perceived challenge level of the task trial by selecting one of the following categories: \emph{easy}, \emph{moderate}, and \emph{hard}. Then, excluding initialization, the participant compares the perceived challenge level of the task trial with that of the previous iteration. Based on the modeled probabilities of the responses, the qualitative GP regression model is updated.

The range of assistance level is represented by $A=\{ x \subset \mathbb{R}^d : 0 \le x_i \le 1\}$, as in the quantitative GP regression model. Let $f_{\mathcal{Q}}(x)$ be the latent function representing the participant's perceived challenge level of the task. The prior distribution of $f_{\mathcal{Q}}(x)$ is modeled with a normal GP distribution. \vspace{-3mm}

\begin{equation}\label{prior f PDF Eq}
f_{\mathcal{Q}}(\boldsymbol{x}) \sim \mathcal{GP}(0,K_{\mathcal{Q}})
\end{equation} 

\noindent where $K_{\mathcal{Q}}\in \mathbb{R}^{n\times n},k_{\mathcal{Q}_{i,j}}= k_{\mathcal{Q}}(x_i,x_j)$ is the noiseless kernel matrix of the qualitative GP regression model. It is worth noting that the kernel functions used in both models need not be identical.  

Let $\boldsymbol{q}$ be the set of qualitative feedback provided by the participant, where $\boldsymbol{q}$ consists of both ordinal classifications $q_o=\{q_{o_1},q_{o_1},...,q_{o_n}\}$ and pairwise preferences $q_p=\{q_{p_2},q_{p_3},...,q_{p_n}\}$. The dataset including ordinal classifications is defined as $\boldsymbol{D_{\mathcal{O}}}=\{(x_1,q_{o1}),(x_2,q_{o2}),...,(x_n,q_{on})\}$, and the dataset including pairwise comparisons is defined as $\boldsymbol{D_{\mathcal{P}}}=\{(x_1,x_2,q_{p_1}),(x_2,x_3,q_{p_2}),...,(x_{n-1},x_n,q_{p_n})\}$. Then, the dataset consisting of all qualitative feedback can be defined as $\boldsymbol{D_{\mathcal{Q}}}=\boldsymbol{D_{\mathcal{O}}}\cup \boldsymbol{D_{\mathcal{P}}}$.

The probability of the latent function based on the provided feedback is calculated from the proportionality $P(f_{\mathcal{Q}}|\boldsymbol{D_{\mathcal{Q}}})\propto P(\boldsymbol{D_{\mathcal{Q}}}|f_{\mathcal{Q}})P(f_{\mathcal{Q}})$, where $P(f_{\mathcal{Q}})$ is the prior unbiased probability of the regression model and $P(\boldsymbol{D_{\mathcal{Q}}}|f_{\mathcal{Q}})$ is the probability of the qualitative feedback of the participant is correct based on the given latent function.

Ordinal classifications and pairwise comparisons are modeled as in~\cite{Biyik2020, Chu2005, Li2020, Catkin2023, Tolasa2024}, as follows: Let $O=\{o_1,o_2,o_3\}$ be the finite set of three ordinal classifications representing the perceived challenge level from ``easy" to ``hard". Let $t$ be the set of ordered thresholds used to distinguish ordinal classifications $t=\{t_{o_0},t_{o_1},t_{o_2},t_{o_3}\}$ and $-\infty=t_{o_0}<t_{o_1}<t_{o_2}<t_{o_3}=\infty$. Then, the probability for the participant to correctly classify a parameter $x_i$ with the ordinal class $o_j^{th}$ is modeled as \vspace{-3mm}

{\small
\begin{equation}\label{Ordinal Class Prob Eq}\!\!\! P(q_{o_i}\!=\!o_j|f_{\mathcal{Q}})\!=\! \Phi\!\left(\!\frac{t_{o_j}-f_{\mathcal{Q}}(x_i))}{c_o}\!\right)\!\!-\Phi\!\left(\!\frac{t_{o{j-1}}-f_{\mathcal{Q}}(x_i)}{c_o}\!\right)\end{equation}} \vspace{-3mm}

\normalsize
\noindent where $\Phi$ represents the cumulative distribution function of the Gaussian distribution and $c_o>0$ is used to capture the classification noise.

The probability of the participant correctly identifying the harder one between two task trials is given by \vspace{-3mm}

{\small
\begin{equation}\label{eqn: Pairwise Comp Prob Eq}\!\!\! P(q_{p_i}=(x_{i}\succ x_{i-1})|f_{\mathcal{Q}})=\Phi\left(\frac{(f_{\mathcal{Q}}(x_i)\ )-f_{\mathcal{Q}}(x_{i-1})\ )}{c_p }\right)
\end{equation} 
} \normalsize \vspace{-3mm}

\noindent where $c_p>0$ captures the noise in pair-wise preferences.

Under the assumption of the independence of provided qualitative feedback, $P(\boldsymbol{D_{\mathcal{Q}}}|f_{\mathcal{Q}})=P(\boldsymbol{D_{\mathcal{O}}}|f_{\mathcal{Q}})P(\boldsymbol{D_{\mathcal{P}}}|f_{\mathcal{Q}})$ is calculated as \vspace{-3mm}

\begin{equation}\label{eqn: Qualitative Dataset Probability}P(\boldsymbol{D_{\mathcal{Q}}}|f_{\mathcal{Q}})=\prod_{i=1}^{n}P(q_{o_i}|f_{\mathcal{Q}})\prod_{i=2}^{n}P(q_{p_i}|f_{\mathcal{Q}}).
\end{equation}
\vspace{-3mm}

The posterior distribution of the regression model is computed by the Laplace approximation~\cite{Rasmussen2005}. Utilizing this commonly adopted method for posterior approximation, it becomes possible to make predictions for Bayesian optimization and extend the probabilistic derivations to capture qualitative feedback from humans~\cite{Biyik2020, Tucker2020, Li2020, Tolasa2024}. 

The perceived challenge level estimate for the participant $f_{\mathcal{Q}_*|\boldsymbol{D_{\mathcal{Q}}}}$ for any arbitrary assistance level $x_*$ can be computed using the posterior model as follows:
\vspace{-4mm}

\begin{eqnarray}\label{eqn: Inference Distribution}
f_{\mathcal{Q}_*|\boldsymbol{D_{\mathcal{Q}}}} \!\!\!&\sim& \!\!\!GP(\mu_{*|\boldsymbol{D_{\mathcal{Q}}}},\sigma_{*|\boldsymbol{D_{\mathcal{Q}}}}^2) \\
\mu_{*|\boldsymbol{D_{\mathcal{Q}}}} \!\!\! &=&\!\!\!  k_{\mathcal{Q}_{*,1:n}} \, K_{\mathcal{Q}}^{-1}\hat{f}_{\mathcal{Q}}  \label{Inference Expectation}  \\
\sigma_{*|\boldsymbol{D_{\mathcal{Q}}}}^2 \!\!\!&=&\!\!\!  k_{\mathcal{Q}_{**}}-k_{\mathcal{Q}_{*,1:n}} \, (W^{-1}+K_{\mathcal{Q}})^{-1}k_{\mathcal{Q}_{1:n,*}} \label{Inference Variation} 
\end{eqnarray} 

\noindent where $K$ is the noiseless covariance matrix calculated with RBF kernel, $W$ is the negative Hessian matrix

\begin{equation}\label{Negative Hessian}W_{ij}=-\frac{\partial^2 \, log(P(\boldsymbol{D_{\mathcal{Q}}}|f(x))}{\partial f(x_i) \, \partial f(x_j)} \nonumber\end{equation} 

\noindent and $\hat{f}$ is the latent function that maximizes the log-likelihood 
\begin{equation}\nonumber
 \hat{f}=argmax_{f(x)}\left(log(P(\boldsymbol{D_{\mathcal{Q}}}|f_{\mathcal{Q}})P(f_{\mathcal{Q}}))\right).
\noindent\end{equation}

\subsection{Sampling Strategy for HiL Pareto Characterization} \label{Sec:SamplingPareto}

Let $\alpha_{num}$ be the acquisition function of $GP_{num}$ and let  $\alpha_{qual}$ be the acquisition function of $GP_{qual}$.
Given these acquisition functions, first, a computationally cheap Pareto optimization problem is solved to find the parameter set $x_p$ lying on the surrogate Pareto front. Next, a parameter is selected among $x_p$ based on the maximum volumetric uncertainty, where the volumetric uncertainty is calculated by multiplying the standard deviations as follows: \vspace{-4mm}

\begin{eqnarray} \noindent
x_p &=&\mathrm{argmax_{x \in {A}}} \, (\alpha_{num},\alpha_{qual})
\\
x_n &=& \mathrm{argmax_{\,x \in x_p}} \,(\sigma_{(x)|\boldsymbol{D_{\mathcal{N}_{n-1}}}} \!\times\sigma_{(x)|\boldsymbol{D_{\mathcal{Q}_{n-1}}}}).
 \noindent\end{eqnarray}

\subsection{Acquisition Function and Hyper-Parameter Selection}\label{Sec:HyperparameterSelection}

Acquisition functions and hyperparameters of the HiL optimization utilized in this study are empirically determined, based on expertise gained through pilot experiments. Ten trials were observed to be sufficient for HiL Pareto characterizations.

In particular, a radial basis function~(RBF) kernel, $k_{{i,j}}~=~\exp(-\theta{||x_i-x_j||}_2^2)$, is used for modeling both $GP_{num}$ and $GP_{qual}$. The kernel hyperparameters $\theta_{\mathcal{N}}$ for $GP_{num}$ and $\theta_{\mathcal{Q}}$ for $GP_{qual}$ are both selected as $5$. The observation noise hyper-parameter $\sigma^2_w$ of $GP_{num}$ is selected as $0.1$. The hyperparameters $c_p$ and $c_o$, capturing the pairwise preference and ordinal classification noise, are selected as $0.5$ and $1$, respectively. The ordinal classification thresholds are selected as $t=\{-\infty, -0.5,0.5,\infty\}$.

For the parameter sampling strategy, two upper confidence bound~(UCB) acquisition functions are used. To sample an assistance level $x_{n}$ for ${n}^{th}$ trial, $\alpha_{num}$ and $\alpha_{qual}$ are used as \vspace{-3mm}

 \begin{eqnarray}
 \alpha_{num}(x_*)_n=\mu_{*|\boldsymbol{D_{\mathcal{N}_{n-1}}}}+\lambda_{\mathcal{N}} \,  \sigma_{*|\boldsymbol{D_{\mathcal{N}_{n-1}}}} 
 \\
  \alpha_{qual}(x_*)_n=\mu_{*|\boldsymbol{D_{\mathcal{Q}_{n-1}}}}+\lambda_{\mathcal{Q}} \, \sigma_{*|\boldsymbol{D_{\mathcal{Q}_{n-1}}}}
\end{eqnarray}
 
 \noindent where $\lambda_{\mathcal{N}}$ and $\lambda_{\mathcal{Q}}$ are the hyper-parameters of $GP_{num}$ and $GP_{qual}$, respectively, and $x_*$ is an any arbitrary feasible assistance level. In these equations, $\lambda_{\mathcal{N}}$ is selected as $2$, while $\lambda_{\mathcal{Q}}$ is selected as $1$.

Once the human-subject experiments were completed, we compared the empirically selected GP hyperparameters with the MLE-optimized values determined from the data and verified that the Pareto-optimal assistance levels obtained under both parameter sets remain close to each other.

\vspace{-3mm}
\subsection{Sample Design Selection for Pareto-based AAN Training} \label{Sec:selection}

All Pareto solutions characterizing the trade-off between the user performance and the perceived challenge are optimal.  Without additional preference information, all Pareto optimal solutions are mathematically equal. The Pareto selection process imposes additional preferences on the set of Pareto optimal solutions to select a subset among them. Pareto optimization methods allow the designer to impose additional constraints \emph{after} the computation of the non-dominated solutions and inspection of the trade-off involved among the conflicting objectives. Unlike scalarization approaches, where the preferences need to be assigned a priori, the ability to impose additional constraints on the set of optimal solutions corresponding to different preferences in a \emph{multi-shot} manner is among the most beneficial aspects of Pareto approaches.

\begin{figure}[h!]  \vspace{-.25\baselineskip}
\centerline{\includegraphics[width=.375\textwidth]{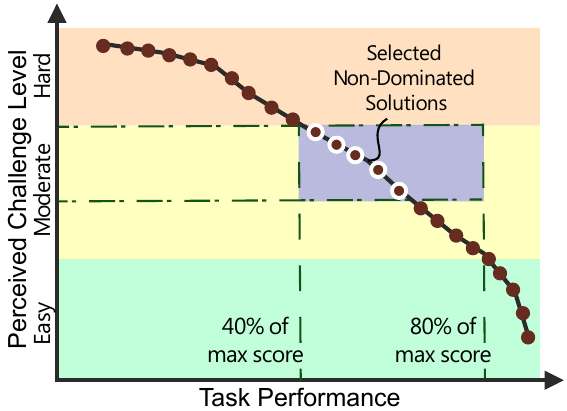}}
\vspace{-.75\baselineskip}
\caption{Selection of non-dominated solutions from the Pareto-front by introducing design constraints after characterizing the trade-off} 

\label{fig:design_selection}
\end{figure}

The trade-off between performance and task difficulty characterized via HiL Pareto optimization can provide insights to guide the design of AAN algorithms. To demonstrate how such protocols can be designed by the characterized trade-off, we propose and test the efficacy of a \emph{sample} AAN training protocol. By imposing new constraints on the Pareto-front based on domain expertise, we form the AAN training protocol as shown in Figure~\ref{fig:design_selection}, so that the controller keeps the task moderately challenging while allowing the user to achieve an acceptable level of performance.

In particular, to facilitate a design selection among all non-dominated solutions, we introduce \emph{personalized} thresholds to the set of non-dominated solutions to keep the perceived challenge level and quantitative performance sufficiently high, as decided by the domain expert. In the sample protocol, the performance range is limited to 40--80\% of the capacity of the individual, while the challenge range is limited to 40--80\% of the perceived challenge level of the individual. All solutions in this range have been selected to be included in the training session, to induce diversity in the training exercises.  The proposed AAN training method emphasizes the importance of customization of the assistance based on individual performance characterizations.

It is important to note that the resulting AAN training protocol based on HiL Pareto characterization is merely a \emph{sample} design selected by the domain expert that utilizes a set of non-dominated solutions for diversity-focused personalized AAN training. The Pareto selection process is not unique, and alternative training methods may be devised from the same set of Pareto solutions. While each of these protocols is equally valid from a multi-criteria design perspective, their training efficacy is likely to vary significantly. 

To examine how the selected threshold affects the set of assistance levels used during AAN training, we evaluated alternative thresholds on the Pareto solutions after the human-subject experiments were completed. The analysis reveals that threshold ranges of 30--70\%, 40--80\% and 50--90\% yield mean assistance levels of $30.6 \pm 22.3$\%, $39.3 \pm 22.6$\%, and $52.3 \pm 17.8$\%, respectively. As expected, shifting the thresholds to cover higher performance yields increased assistance levels.

Given different preferences yield different AAN strategies with varying training outcomes, more sophisticated designs, such as an adaptive protocol for selecting the appropriate threshold ranges to be imposed on Pareto solutions to adjust the assistance levels based on instantaneous user performance on the fly, may perform better than the sample AAN protocol in terms of training efficacy. The sample AAN protocol is preferred due to its simplicity, as the main goal of this study is to illustrate the feasibility of such designs based on HiL Pareto characterizations.

\section{Human Subject Experiment}

This section presents the application of the proposed HiL Pareto optimization approach to a manual skill training task with haptic feedback, administered to healthy volunteers. It demonstrates that HiL Pareto optimization can be used to (i)~efficiently and systematically characterize the trade-off between the task performance and the perceived challenge level, (ii)~design AAN training protocols, and (iii)~establish a novel and rigorous means of performance evaluation under assistance, within and between participants and groups.

\vspace{-3mm}
\subsection{Participants}
{34}~participants, with an average age of {23.6 $\pm$ 3.6} years participated in the experiment. The participants performed a motor learning task using a force-feedback joystick with their dominant hand. All participants signed the informed consent form approved by the Institutional Review Board of Sabanci University~{(Protocol No:~FENS-2025-18)}.

Participants with any known sensory-motor disability or significant prior experience with haptic devices are excluded from the study. Similarly, ceiling pre-training performance of 80\% under less than 40\% assistance was used as an exclusion criterion to avoid saturation during training. Accordingly, participants completed warm-up and pre-evaluation sessions to verify that the task is sufficiently challenging; two participants who achieved ceiling-level performance were excluded from the study before the group assignments. Eligible participants were assigned to experimental groups using a Latin square method to achieve balanced group allocations. Participants were blinded to group identity and protocol differences throughout the experiment. All assigned participants completed the full protocol without technical failures or withdrawal, resulting in no attrition or missing data.

\vspace{-1mm}
\subsection{Task and Apparatus}
\begin{figure}[h!] \vspace{-.45\baselineskip}
\centerline{\includegraphics[width=.425\textwidth]{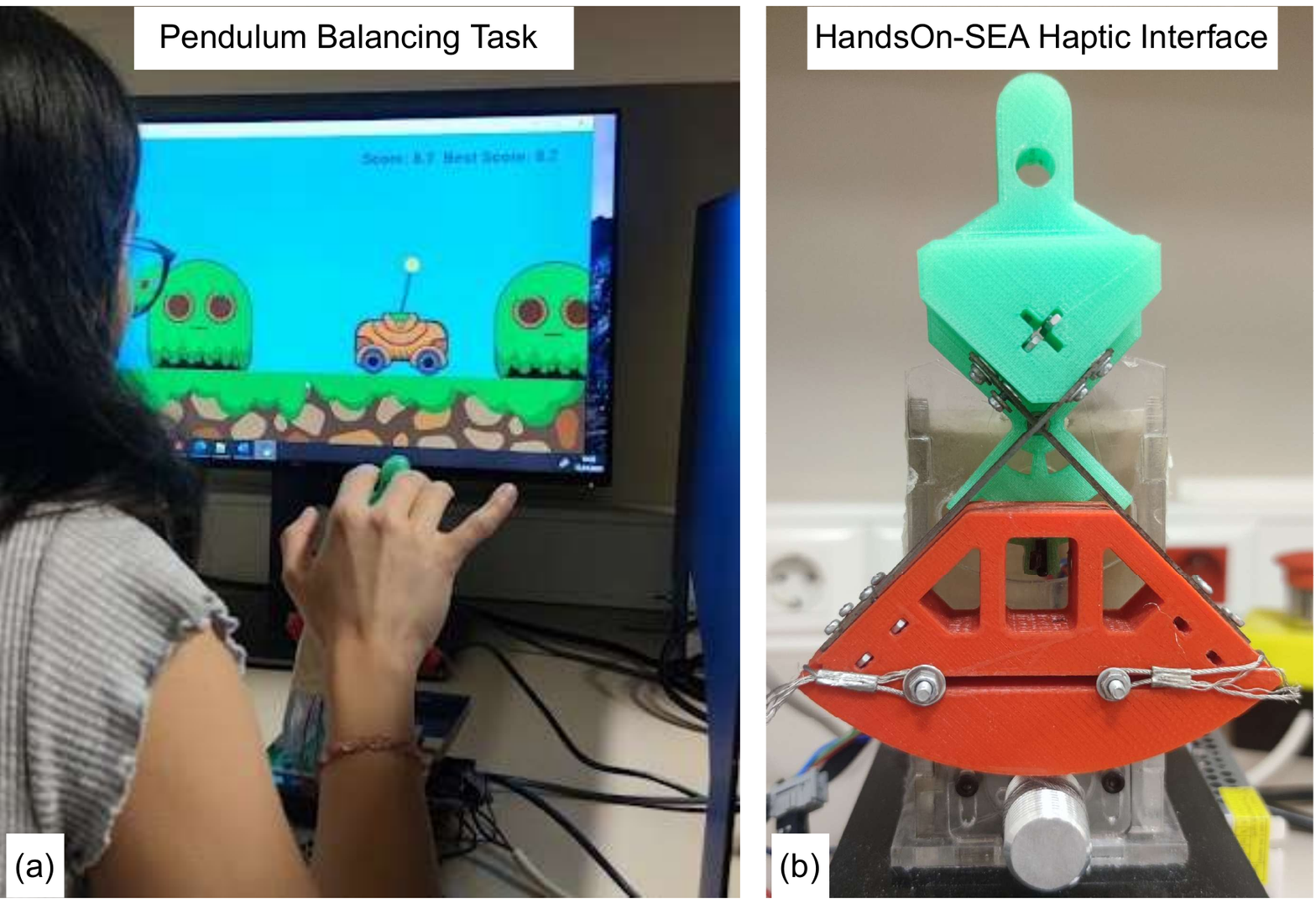}}
\vspace{-4mm}
\caption{(a)~A participant holding a force-feedback joystick~(HandsOn-SEA) while interacting with the pendulum game, (b)~HandsOn-SEA haptic interface}
\label{fig:game}
\end{figure}

The motor learning task consists of a two-dimensional balancing of a virtual inverted pendulum on a cart, displayed on an LCD screen, as shown in Figure~\ref{fig:game}(a). A single-axis force-feedback joystick, called HandsOn-SEA, presented in Figure~\ref{fig:game}(b), renders the dynamics of the cart and pendulum model, so that users feel as if they are moving the cart while operating the joystick. Assistance forces are also provided through this interface. Two monsters, separated by a constant distance, provide continuous disturbances. The external disturbances are generated using a stochastic process with identical parameters across participants and trials. The task fails if the pendulum rotates more than $\pm$50$^{\circ}$ or the cart touches the monsters. The objective of the game is to survive for 25~s, with raw scores corresponding to the survival time and normalized scores divided by 25~s. 

The balancing task is intentionally designed to be challenging to avoid performance saturation; however, due to the high sensitivity of the task dynamics, brief lapses of attention can easily lead to failures. Accordingly, participants are provided with three chances per trial, and a best-of-three score is used as a more robust and optimistic performance metric to reduce noise and avoid under-evaluating participant performance due to such momentary distractions.

\begin{figure*}[t!]
\centering 
\includegraphics[width=.9\textwidth]{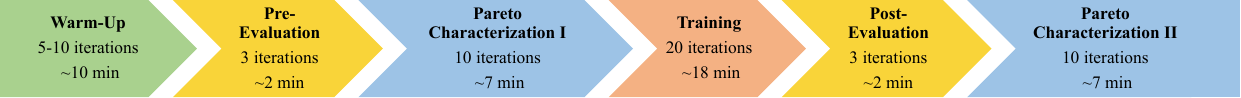}
\vspace{-3mm}
\caption{Experimental procedure} 
\vspace{-.85\baselineskip}
\label{fig:experimentalProcedure}
\end{figure*}

Interaction and assistance forces are rendered through HandsOn-SEA, a custom-built single-axis haptic interface with series elastic actuation~(SEA)~\cite{Otaran}. HandsOn-SEA features a coreless DC motor equipped with an encoder to actuate its sector pulley through a capstan-drive to impose desired motions. A cross-flexure pivot, formed by crossing two symmetric leaf springs, acts as the compliant element located between the rigid sector pulley and the handle structures. A Hall-effect sensor constrained to move between the neodymium block magnets embedded in the sector pulley measures the deflections of the compliant element, thereby enabling estimation of the interaction force. HandsOn-SEA can provide 15~N continuous force output at its handle with a force control bandwidth of 12~Hz, within a workspace of $\pm55^{\circ}$. HandsOn-SEA works under velocity-sourced impedance control~\cite{FatihEmre2020, Kenanoglu2024, Kenanoglu2025a, Kenanoglu2025b}, implemented in real-time at 1~kHz, utilizing a TI~C2000 microcontroller. The microcontroller communicates with the host computer, displaying the game via serial communication.

To determine assistance forces, an LQR controller is implemented to determine the appropriate cart positions to stabilize the pendulum and to move the cart to the middle of the monsters. Along with the cart and pendulum dynamics, a virtual spring is rendered between the stabilizing controller-determined ideal position and the current position of the cart by HandsOn-SEA to provide assistance forces to participants. The assistance is adjusted through the stiffness of the coupling spring; the larger the spring constant, the larger the~assistance. 

\vspace{-2.5mm}
\subsection{Experimental Conditions}

The training efficacy of the proposed AAN training method based on Pareto optimization (test group) is compared with a control group
based on a commonly employed performance-based adaptive assistance controller. Each group had an equal number of participants. The following protocols are compared:
\begin{itemize}
\item[i)]{\emph{Test Group -- Pareto Approach:}} As detailed in Section~\ref{Sec:selection}, the AAN training method based on Pareto optimization relies on HiL characterizations of the trade-off between the task performance and the perceived change level. Pareto-based approach naturally results in a diverse set of assistance levels. After a trade-off is characterized, thresholds are introduced to select the subset of non-dominated solutions that span from 40\% to 80\%  of the performance and the perceived challenge level of individual users. Non-dominated solutions in this subset are used in the AAN training session by selecting the assistance levels in a randomized order for each trial. If the number of non-dominated solutions is less than the number of trials, then the random selection process is restarted from the original non-dominated solution subset, after all solutions in the subset are used. On average, $32.9 \pm 24.0$ non-dominated solutions were located within the 40--80\% thresholds. \smallskip

\item[ii)]{\emph{Control Group -- Adaptive Assistance:}} The control group is trained with a commonly used adaptive assistance controller. A performance-based adaptive assistance controller is implemented using an adaptive staircase approach, as in~\cite{Yanfang2009, Anguera2013, Gray2017}. The adaptive approach is an online local search method that seeks a single assistance level for a participant during trials. The assistance level of the control group is initialized at $50\%$ and employs a two-up-one-down variation. In this adaptive approach, the assistance decreases by $10\%$ after every two consecutive successful trials or increases by $10\%$ after each failure.
\end{itemize}

\vspace{-3mm}
\subsection{Experimental Procedure}

The experiment consists of 6 sessions, as in Figure~\ref{fig:experimentalProcedure}: warm-up, pre-evaluation, pre-training HiL characterization, training, post-evaluation, and post-training HiL characterization. The experimental procedure was administered identically for the control and the test groups, except for the assistance levels provided to each group during the training session.

The warm-up session includes a tutorial to help participants become familiar with the rules and mechanics of the game. Participants play the game with several levels of assistance. 
The warm-up session is concluded when a participant displays a clear understanding of the rules and can play the game without unpremeditated failure. 

Pre- and post-evaluations measure the performance of the participants under no assistance. A comparison of pre- versus post-evaluation sessions provides a measure of participants' progress after the training sessions.

HiL Pareto characterization sessions are used to learn the trade-off between a participant's quantitative performance and the perceived challenge level of the game. For this purpose, two GP regression models are trained in each HiL session. During each iteration of the HiL learning, the participant plays the game with an assistance level determined by the optimization algorithm. Once the game ends, the best score of the participant is used to train a quantitative GP model. 

After each game, two questions are asked to the participant to determine the perceived difficulty of the game. First, the participant is asked to rate the challenge level of the game by selecting one option from ``hard", ``moderate", or ``easy", resulting in an ordinal classification. Next, (except for the first trial), the participant is asked to compare the challenge level of the last game with respect to the previous one in a pair-wise manner. The answers to these queries are used to train a qualitative GP regression model. Finally, utilizing a multi-criteria sampling strategy, the HiL optimization algorithm updates the assistance level provided for the next iteration.

\begin{figure*}[t!]  
\centerline{\includegraphics[width= \textwidth]{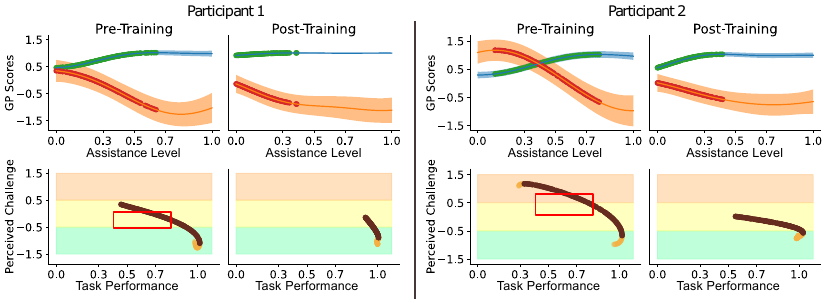}}
 \vspace{-.85\baselineskip}
\caption{In the top rows, the orange and blue lines show the mean of the surrogate function for the perceived challenge and the task performance, respectively, while the shading depicts their standard deviation. The bottom rows present the Pareto solutions with dark red dots, while the dominated solutions in the feasible set are shown with orange dots. The Pareto solutions inside the red squares are used during the training of the participants in the test group.}
\label{fig:paretos}
\vspace{-0.25\baselineskip}
\end{figure*}

\vspace{-1.5mm}
\subsection{Hypotheses} \label{Sec:hypothesis}
The human subject experiments aim to test the validity of the following hypotheses: \smallskip

\begin{itemize}
\item[$H_1$] The trade-off between the performance and the perceived challenge level of a task can be characterized by utilizing a HiL Pareto optimization approach.
\item[$H_2$] Comparisons of the trade-off curves characterized at the different stages of training provide a rigorous means of evaluating training performance within and between participants, across feasible assistance levels.
\item[$H_3$] The non-dominated solutions characterizing the trade-off between the performance and the perceived challenge can be used to design an AAN training protocol. 

\end{itemize}

\section{Results and Discussion} \label{Sec:results}

In this section, we elaborate on each hypothesis presented in Section~\ref{Sec:hypothesis} in the view of the experimental evidence. We also demonstrate how the HiL Pareto optimization approach can be used to assess group-level progress under all feasible assistance levels.
\subsection{Hypothesis~1}

Figure~\ref{fig:paretos} presents the GP regression models and the Pareto fronts, before and after the training, for two sample participants with high and intermediate performance. {The data collected during the experiments are available for download from our laboratory website\footnote{{\url{http://hmi.sabanciuniv.edu/HiL\_Pareto\_Experiment\_Data.zip}}}}. In the top rows, the orange lines show the mean of the surrogate function for the perceived challenge level, while the orange shading depicts its standard deviation. Similarly, the blue lines represent the mean prediction for the game scores, while the blue shading depicts their standard deviation. The bottom rows present the Pareto plots characterized during pre- and post-training sessions, shown in the left and right columns. 

In the Pareto plots, the black points represent Pareto solutions, while the orange points depict all feasible solutions for the problem. Pareto solutions cover a non-trivial (and possibly disconnected, as in Figure~\ref{fig:pareto2}) subset of the set of all feasible solutions, as they capture the non-dominated solutions of the trade-off between the expectations of the perceived challenge level and the task performance (game score). For ease of visualization, the Pareto plots are divided into three regions; the red, yellow, and green shades represent the perceived challenge levels of hard, moderate, and easy, respectively. Finally, the red squares indicate the Pareto solutions used during the training of the participants in the test group, according to the Pareto selection criteria detailed in Section~\ref{Sec:selection}.

As hypothesized in $H_1$, Figure~\ref{fig:paretos} provides evidence that the trade-off between the performance and the perceived challenge level of a task can be characterized via HiL Pareto optimization with hybrid performance measures. Thanks to the sample-efficiency inherited from the underlying  Bayesian optimization approach~\cite{Belakaria2020}, HiL Pareto characterization converged in {$6.3 \pm 1.8$~min}; hence, it can easily be performed before the training sessions without inducing fatigue to the participants. 

The top rows in Figure~\ref{fig:paretos} verify that the Bayesian multi-criteria optimization technique samples from the regions of surrogate functions that conflict with each other. The efficiency of the sampling strategy in locating the conflicting regions and the uniformity of samples in these regions are key aspects of the efficiency of the Pareto optimization technique that make its use adequate for HiL optimization. 

The bottom rows in Figure~\ref{fig:paretos} characterize the trade-off between the performance and the perceived challenge level of the task.  In particular, in the sample Pareto plots, both participants can achieve their best scores when their perceived challenge level is low, and their performance decreases as their perceived challenge level increases. Furthermore, the trade-off curves also enable the evaluation of whether the task is too challenging or too easy for the participant, by checking their performance under high and low assistance levels, respectively. If such a case is detected, it may be preferable to change the task to better cater to the abilities of the participant.

Overall, the experimental results provide strong evidence that the trade-off between the performance and the perceived challenge level of a task can be characterized by utilizing a HiL Pareto optimization approach. Consequently, the results support the first hypothesis~$H_1$.

\subsection{Hypothesis~2} \label{Hyp3}
 
The proposed Pareto approach emphasizes the customization of training sessions based on individual performances.  Accordingly, this section focuses on \emph{individual-level} results.

In the bottom rows of Figure~\ref{fig:paretos}, the changes between the pre- and post-training trade-off characterization plots capture the progression of the participants under assistance. For Participants~1 and~2, the shifts in the Pareto plots explicitly indicate that participants not only perceive the game as less challenging after the training, but also their scores improve significantly. Accordingly, their post-training Pareto curves have shifted towards the right and downwards, compared to their pre-training Pareto curves. For instance, for Participant~2, the perceived level of challenge for the most difficult trial has reduced from hard to moderate, indicating that the task became easier to perform as learning took place. The positive progress in the performance can also be observed by the shrinkage of the Pareto front towards the right side, which captures the region for higher game scores. This shift indicates the game scores for the most difficult perceived challenge level have increased from 0.30 to 0.55 for Participant~2. 

These improvements can also be observed in the GP regression models, before and after the training. For instance, one can observe by inspecting the surrogate function for the perceived challenge level and game scores of Participant~2 that the performance and challenge level saturate around 70\% assistance before the training, while this saturation shifts to 45\% assistance after the training. Hence, only the assistance levels from 0\% to 40\% belong to the post-training solutions.

\begin{figure}[b!] 
\centerline{\includegraphics[width= .5\textwidth]{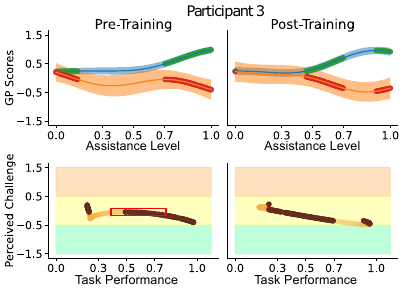}}
 \vspace{-.75\baselineskip}
\caption{Results for a sample participant with low performance. The presentation follows the same format as in Figure~\ref{fig:paretos}.}
\label{fig:pareto2}
\end{figure}

Comparison of the pre- and post-training surrogate functions is especially useful to understand Pareto plots that consist of multiple disconnected sections. Figure~\ref{fig:pareto2} presents the results for a low-performing Participant 3, whose Pareto plots are harder to interpret, but the surrogate functions for the perceived challenge level and the game scores can help understand the results.  While the worst performance of Participant~3 did not significantly improve from pre- to post-training, the shifts in the surrogate functions of this participant indicate that Participant~3 requires less assistance to achieve a similar level of performance to the pre-training case. In particular, the increase in performance and the decrease in challenge level shifts from 70\% assistance in the pre-training characterization results to 50\% assistance in the post-training results. Accordingly, the trade-off characterization captures improvements in the performance that cannot be captured by simply observing the participants' performance without any~assistance. 

Finally, utilizing the HiL Pareto optimization, the trade-off curves of \emph{different participants} can also be compared with each other. For instance, the post-training Pareto front of Participant~2 is slightly better than the pre-training Pareto front of Participant~1, indicating that Participant~2 reaches a more advanced stage after the training, compared to the pre-training performance of Participant~1. Note that rigorous comparisons of different participants with each other, in general, is a very challenging task; a fair comparison of different participants is possible by considering the non-dominated solutions of the Pareto optimization, as they capture the best possible performances for each challenge level~\cite{Yusuf2020,Bonabarxiv2021}.

Overall, the results support that comparisons of the trade-off curves characterized at the different stages of training can provide a rigorous means of evaluating training performance, as hypothesized in $H_2$.

\subsection{Hypothesis~3} \label{Hypothesis1}

\begin{figure}[h!]
\centerline{\includegraphics[width=.38\textwidth]{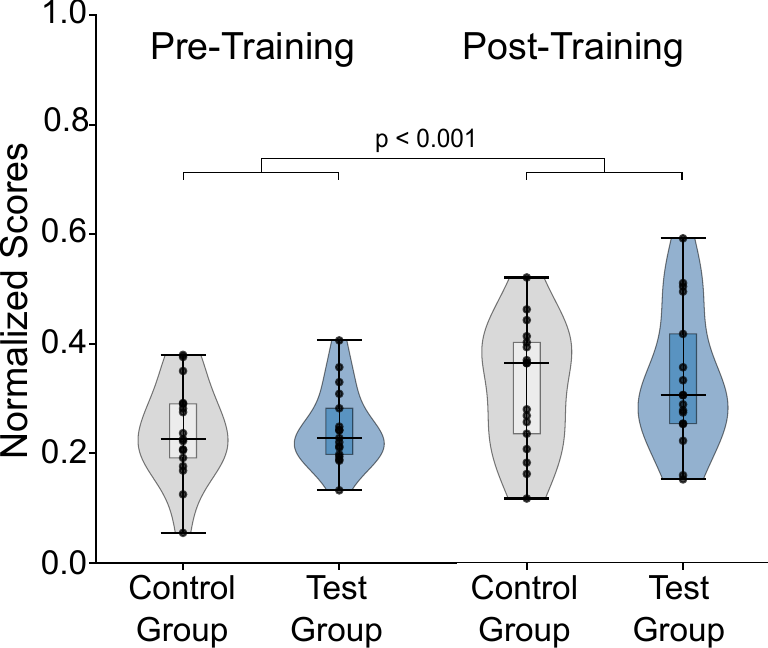}}
\vspace{-.75\baselineskip}
\caption{Violin plots of the unassisted pre- and post-training performance}
\label{fig:boxplot}
\end{figure}

To assess the overall improvement of the participants after the training session, we compare the unassisted game scores of the participants before and after the training sessions, as depicted by violin plots in Figure~\ref{fig:boxplot}. Furthermore, a two-way mixed-design ANOVA is conducted to examine the effects of training phase (pre- vs. post-) and group (test vs. control) on the task performance. Prior to the analysis, assumptions for ANOVA were tested and verified as follows: The Shapiro-Wilk test confirms that the residuals were normally distributed (p~=~.420). Levene’s test indicates the homogeneity of variances for both the pre- (p~=~.406) and post-training phases (p = .955), supporting the assumption of equal variance across the groups.

There is a significant main effect of training phase (f(1,32)~=~37.88, p $<$ .001, partial $\eta^2$ = 0.542), indicating a statistically significant improvement in performance from pre- to post-training across all participants. The main effect of the group is not significant (f(1,32) = 0.13, p~=~.725, partial $\eta^2$ = 0.004), suggesting that overall performance did not differ between the test and control groups. Furthermore, the interaction between the group and the training phase is not significant (f(1,32) = 0.08, p~=~.784, partial $\eta^2$ = 0.002), indicating that both groups showed similar improvements from pre- to post-training.


Since no statistically significant differences are observed under no-assistance conditions, within the power and scope of the statistical analysis, there is no evidence that one protocol outperforms the other. Together with the successful implementation of the Pareto-based protocol, these results support hypothesis~$H_3$. Consequently, the results demonstrate that training based on non-dominated solutions can yield AAN protocols.

\begin{figure*}[t!]
\centerline{\includegraphics[width=\textwidth]{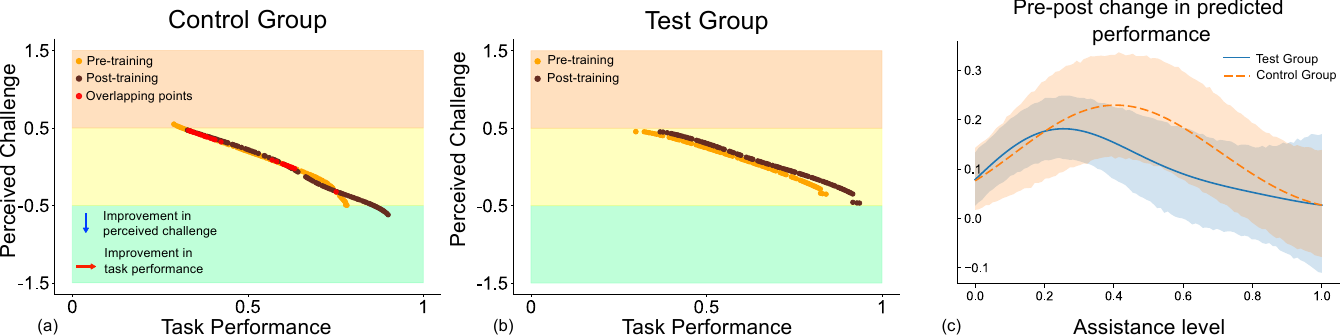}}
\vspace{-.5\baselineskip}
\caption{Comparison of aggregate Pareto solutions of control~(a) and test~(b) groups during pre- and post-training. In these figures, a shift of the Pareto front from left to right indicates improvement in task performance, while a shift from top to bottom indicates the task being perceived as less challenging. (c)~Percent improvement of the available performance of the control and test groups.}
\vspace{-.5\baselineskip}
\label{Fig:aggrate_efficacy}
\end{figure*}

\subsection{Comparison of Group Performance under Assistance}

As discussed in Section~\ref{Introduction},  typical evaluations of manual skill training are performed when no assistance is provided, as in Section~\ref{Hypothesis1}, since this represents the real-life use case. On the other hand, such evaluations cannot capture performance improvements during the early phases of training or under assistance. One of the important insights provided by the proposed HiL Pareto optimization framework is that trade-offs characterized at different stages of training provide a rigorous means to fairly assess the progress of individuals, under all feasible assistance levels, as discussed in Section~\ref{Hyp3}.

In this section, we demonstrate how performance analyses can be performed at a \emph{group-level}. In particular, we study the average improvements of the control and test groups through their aggregate GP models. The goal of such analyses is not necessarily to show that one group is superior to the other, but to gain further insights into the performance of both training protocols under all feasible assistance levels. 

To estimate group-level performance, we first aggregate trained GP models by statistically averaging them among the participants and use these aggregate GP models to derive aggregate Pareto curves for each group during pre- and post-training characterizations. Figures~\ref{Fig:aggrate_efficacy}(a) and~(b) present aggregate Pareto plots for the control and test groups, respectively. These comparisons across assistance levels and pre/post-training characterization instances are rigorous and fair, as the Pareto solutions capture the best achievable performance predicted for each group under all feasible assistance conditions at each characterization instance. The shifts of the aggregate Pareto curves from pre- to post-training in these figures indicate observable improvements in the average task performance of both groups, together with minor improvements in their average perception of task difficulty.

Next, we investigate the difference in pre- and post-training performance predicted by the aggregate GP models of the two groups. Figure~\ref{Fig:aggrate_efficacy}(c) presents, for each assistance level, the mean change in GP-predicted performance together with 95\% confidence intervals obtained via non-parametric bootstrapping. For each bootstrap replicate, participants are resampled with replacement, the aggregate GP model is evaluated at each assistance level, and the group-level mean change is recomputed. A total of 5000 bootstrap samples are used, and percentile-based confidence intervals are reported.

\begin{figure}[b!]
\centerline{\includegraphics[width=.425\textwidth]{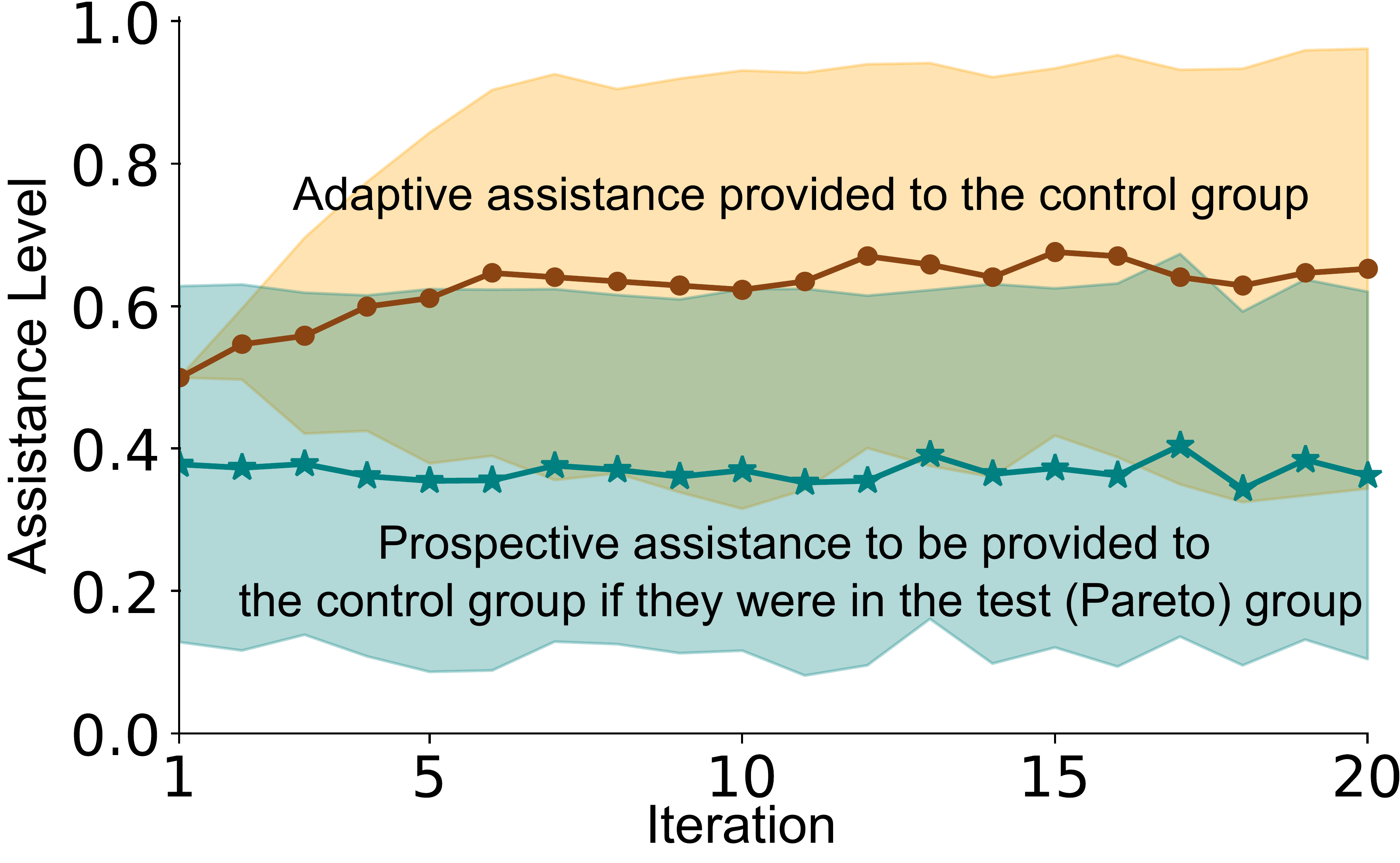}}
\vspace{-.85\baselineskip}
\caption{Assistance provided to participants in the control group (adaptive assistance) versus the \emph{prospective assistance} that would have been provided to these participants in the control group \emph{if they were in the test group} (Pareto approach). Lines show the average assistance levels provided to participants, while shaded areas indicate one standard deviation.}
\label{Fig:assistance_provided}
\end{figure}

Figure~\ref{Fig:aggrate_efficacy}(c) shows that both groups possess wide confidence intervals at each assistance level, reflecting substantial inter-participant variability in GP-predicted performance improvements. Given the confidence intervals for the two groups largely overlap, it can be concluded that the overall performance improvement is broadly similar for both training protocols across assistance levels. As a descriptive trend, the test (Pareto-based)  group achieves slightly higher mean performance gains, especially within the 40\%-90\% assistance level range. While these differences are not statistically robust for this particular human-subject experiment, similar analyses may lead to stronger trends for other experiments, providing valuable insights to the designers for making informed decisions while further optimizing the training protocol.

To have a better understanding of the assistance levels selected by the AAN protocols for an \emph{identical} group of participants, Figure~\ref{Fig:assistance_provided} presents the assistance level provided to the control group in comparison to the \emph{prospective assistance level} that would have been provided to the \emph{same} control group, if these participants were placed in the test group. Since the assistance levels used in the Pareto-based training protocol are independent of the participant's instantaneous performance, they can be determined solely from their corresponding Pareto solutions. Consequently, the prospective assistance levels to be provided to the participants if they were in the test group are determined based on their pre-training HiL characterization.

Figure~\ref{Fig:assistance_provided} indicates that the adaptive staircase method provided mean assistance levels around 60\% to the control group, while the Pareto-based approach \emph{would have} selected prospective assistance levels closer to 40\% on average for the same set of individuals in the control group. The difference between the mean assistance levels of the two approaches provides a possible explanation for larger performance change trends observed for the Pareto-based group between 40\%-90\% assistance levels in Figure~\ref{Fig:aggrate_efficacy}(c). Training under lower mean assistance for the Pareto-based training protocol may have exposed participants to conditions demanding greater active control more frequently, and this could have contributed to the observed trend between 40\%-90\% assistance levels. Since the confidence intervals for both the assistance level distributions in Figure~\ref{Fig:assistance_provided} and the performance changes in Figure~\ref{Fig:aggrate_efficacy} have substantial overlap, the assistance experienced by the two groups is quite similar at the group-level, and the observed trends need to be interpreted with caution. 

Overall, the analysis in this subsection does not provide evidence of significant differences between the two tested training protocols across assistance levels for this particular human-subject experiment. On the other hand, the proposed analysis method, based on comparing aggregate Pareto plots and GP models captured during pre- and post-training, provides a rigorous method to fairly compare the performance of the control and test groups across all assistance levels, thereby enabling the designer to capture potential differences among training protocols to gain useful insights.

\section{Conclusion}

This study demonstrates the feasibility of HiL Pareto optimization, its potential to help with the design of new AAN controllers, and its novel use for individual and group-level performance evaluations and comparisons. Human subject experiments with qualitative and quantitative cost functions are provided to demonstrate the use of two different forms of feedback in HiL Pareto optimization.

While the simplest models are used to promote ease of presentation, the proposed HiL Pareto optimization approach is generic and can be easily extended to Pareto solutions for any number and type of cost functions. Similarly, while this study has been conducted for the single decision variable of the assistance level, the proposed Pareto optimization method trivially extends to a larger number of decision variables. Furthermore, while we have utilized USeMO as an efficient optimizer, similar results may be achieved by utilizing other sample-efficient Pareto optimization approaches. 

\subsection{Limitations of the Study}

While our results show that the Pareto-based training protocol performs comparably with a commonly used adaptive method, the number of participants in this study is sufficient to detect medium effect sizes. As a result, smaller differences between training strategies may have gone undetected. Similarly, group-level comparisons based on GP estimates provide only weak trends for our study, as the confidence regions of groups largely overlap. Further experiments with a larger number of participants, multiple training sessions, and long-term retention evaluations may offer deeper insights into performance differences not captured in this feasibility study.

Additionally, our current experiment evaluates the effectiveness of Pareto-based AAN training using a simple protocol to facilitate the presentation of the underlying Pareto-based design concept. The design of an effective training protocol is an involved process that needs consideration of multiple aspects, such as scheduling of breaks and repetitions, that go beyond the determination of the proper level of assistance.

Overall, it is important to re-emphasize that our focus in this feasibility study is to demonstrate that the characterized trade-off can help with the design of promising and effective AAN protocols by providing insights. Designs of more sophisticated AAN protocols and claims of improved training efficacy require further optimizations, such as sensitivity and robustness analyses, which are beyond the scope of this study.

\subsection{Ongoing Works}
The determination and evaluation of more sophisticated AAN approaches are parts of our ongoing work. Exploring and validating these approaches through systematic experimentation will help us better understand how to optimize AAN training using the HiL Pareto optimization framework.

\begin{figure}[b!]  
\centerline{\includegraphics[width= .375\textwidth]{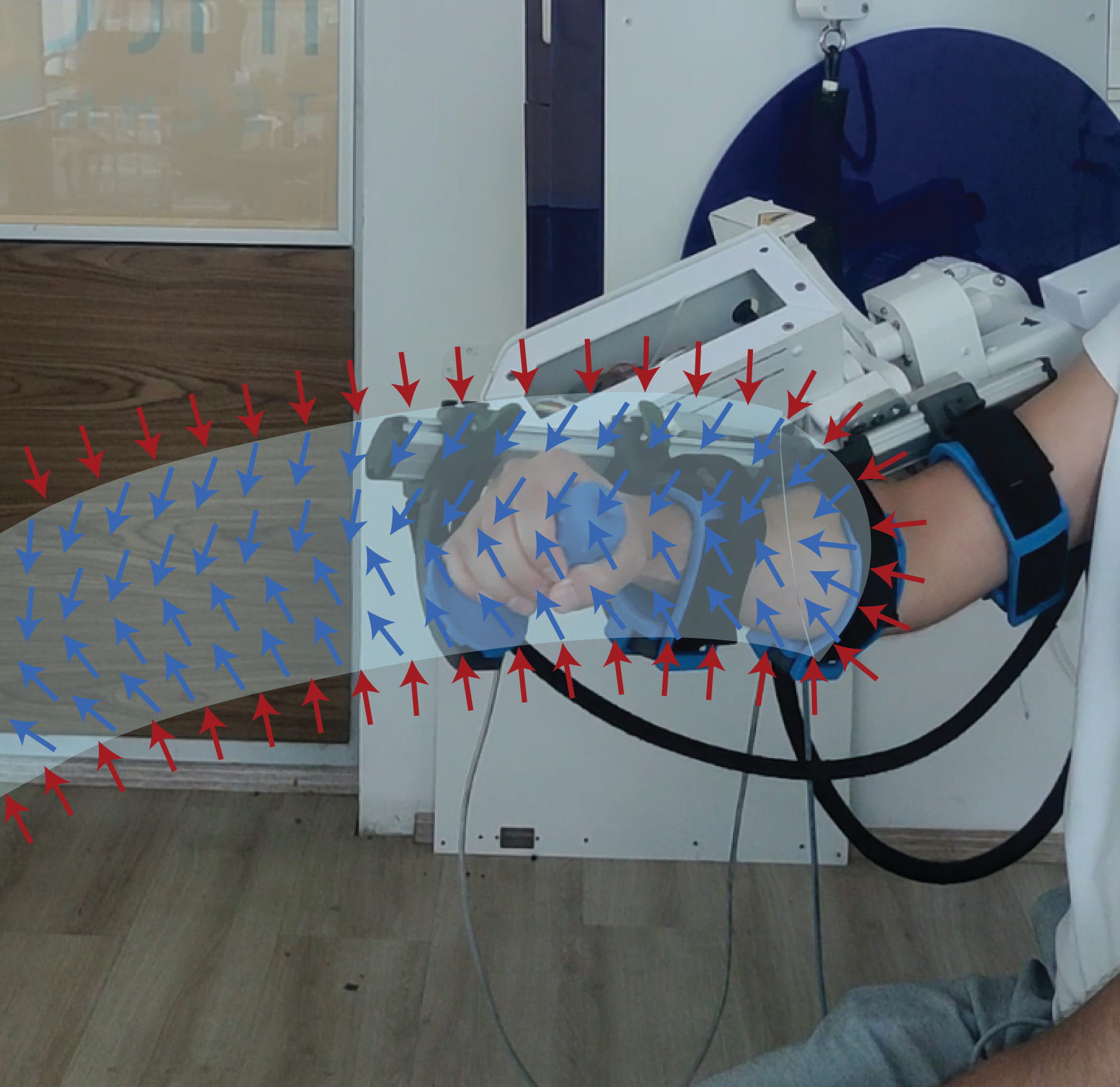}}
 \vspace{-.75\baselineskip}
\caption{A snapshot during HiL Pareto characterization and AAN training with the AssistOn-Arm upper-extremity exoskeleton}
\label{Figure:exo}
\end{figure}

Since the task difficulty-user performance trade-off also exists in the rehabilitation context, the proposed HiL Pareto optimization approach is directly applicable to these applications. Figure~\ref{Figure:exo} provides a snapshot of our ongoing studies, in which HiL Pareto optimization is applied to stroke patients with the AssistOn-Arm six DoF upper extremity exoskeleton~\cite{Argunsah2025,Ergin2012,MustafaVolkan2012}.

For the utilization of the proposed framework in rehabilitation applications, the perceived challenge level cost function and the underlying HiL Pareto optimization method do not require any changes, while the rehabilitation task to be performed and the quantitative performance metric are modified with more clinically relevant ones. HiL Pareto optimization can still be performed over the design variable of the assistance level provided to the user, capturing the force-feedback applied to the patient within a virtual tunnel around the nominal path of the rehabilitation exercise. The multi-DoF nature of the exoskeleton becomes relevant only when the required assistance level in task space is mapped to the joint space of the robot for the appropriate actuator torques. As in the case of motor skill training, the number of design variables can be easily extended to include other relevant variables, such as the diameter of the virtual tunnel utilized during rehabilitation.

Overall, our results indicate that the proposed HiL Pareto optimization approach holds promise for applications in both manual skill training and robot-assisted rehabilitation.

\section*{Acknowledgement} 
This work has been partially supported by TUBITAK Grants~120N523 and~23AG003. 

\bibliographystyle{IEEEtran}
\bibliography{references.bib}

\begin{IEEEbiography}[{\includegraphics[width=1in,height=1.2in,clip,keepaspectratio]{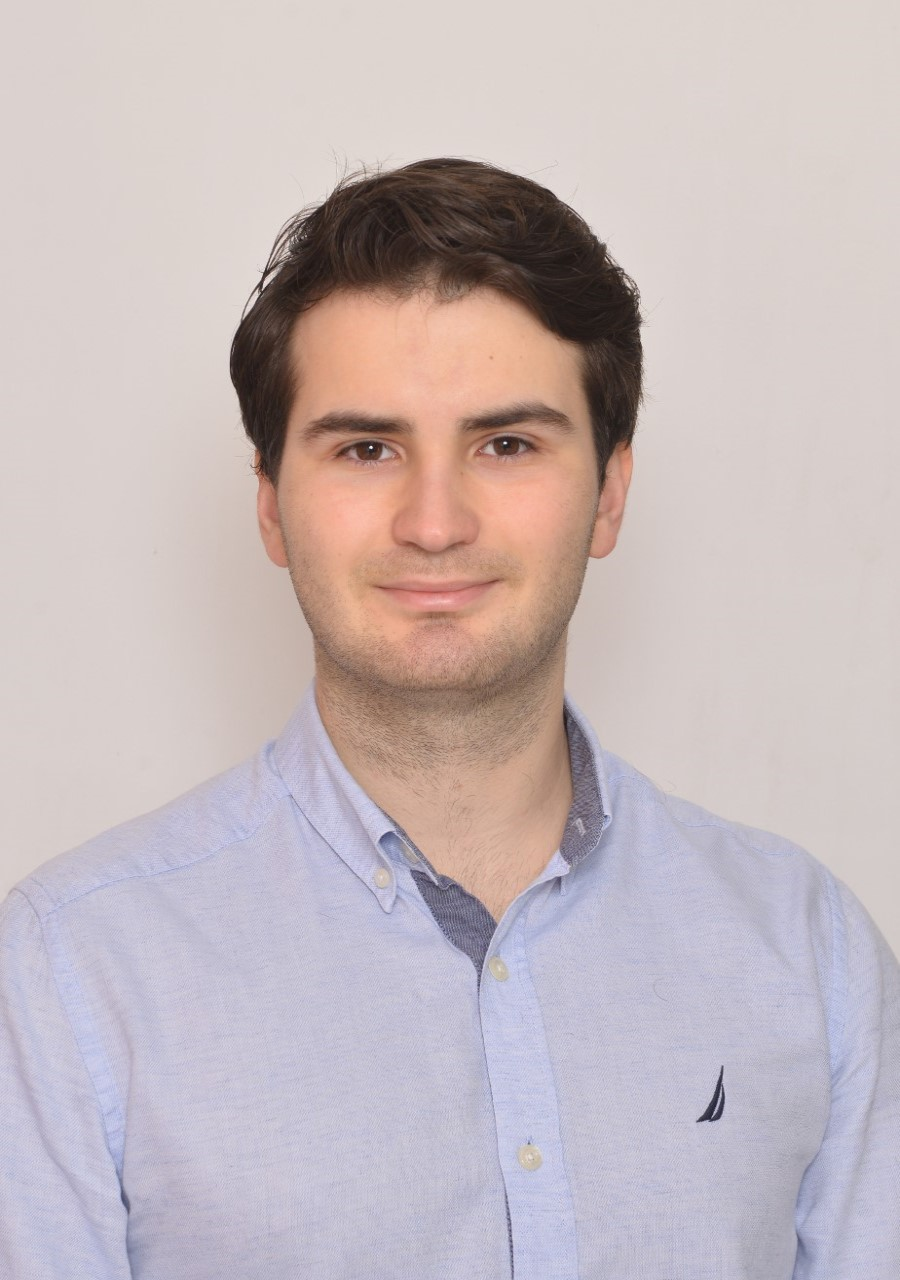}}]
{Harun Tolasa}  received his B.Sc. degree in mechanical engineering from Bilkent University (2021) and his M.Sc. in mechatronics engineering from Sabanci University~(2024). Currently, he is pursuing his Ph.D. degree at Sabanci University. His research interests include active learning, human-in-the-loop optimization, and haptic rendering.
\end{IEEEbiography}

\begin{IEEEbiography}[{\includegraphics[width=1in,height=1.2in,clip,keepaspectratio]{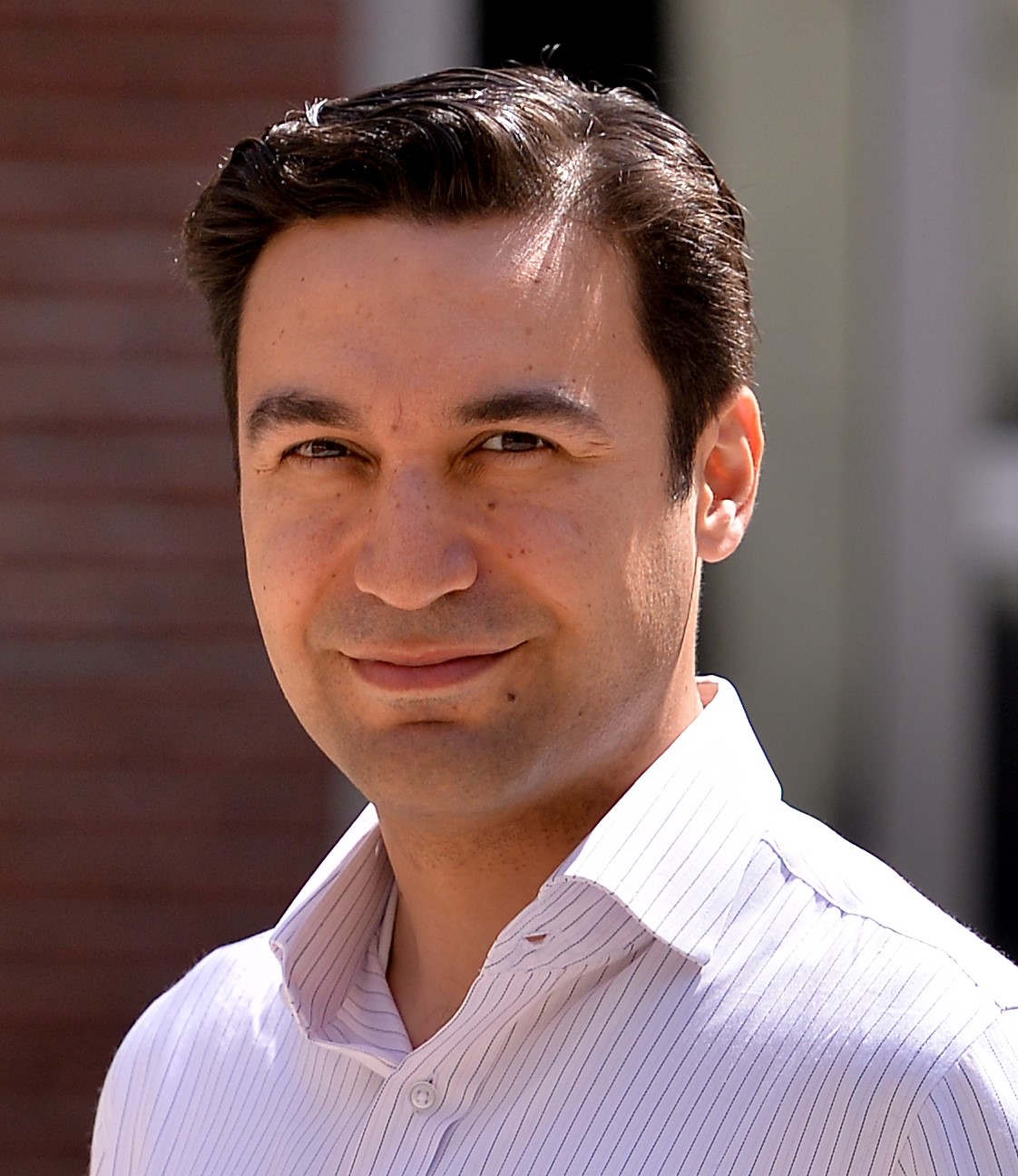}}]
{Volkan Patoglu} is a full professor in mechatronics engineering at Sabanci University.
He received his Ph.D. degree in mechanical engineering from the University of Michigan, Ann Arbor~(2005) and worked as a post-doctoral researcher at Rice University~(2006). His research is in the area of physical human-machine interaction, in particular, design and control of force feedback robotic systems with applications to rehabilitation. His research extends to cognitive robotics. He has served as an associate editor for IEEE Transactions on Haptics~(2013--2017), IEEE Transactions on Neural Systems and Rehabilitation Engineering~(2018--2023), and IEEE Robotics and Automation Letters~(2019--2024).
\end{IEEEbiography}

\vfill

\end{document}